\pdfoutput=1
\documentclass[11pt]{article}
\usepackage{geometry}
\usepackage{coling2020}
\usepackage[utf8]{inputenc}

\usepackage{times}
\usepackage{url}

\usepackage{latexsym}
\usepackage{booktabs}
\usepackage{microtype}
\usepackage{float}
\usepackage{svg}
\usepackage{graphicx}
\usepackage{adjustbox}
\usepackage{enumitem}
\usepackage{subfig}

\usepackage{caption}

\newcommand\AM[1]{}

\newcommand\KM[1]{}
\usepackage{soul}

\colingfinalcopy 

\title{BAKSA at SemEval-2020 Task 9: Bolstering CNN with Self-Attention for Sentiment Analysis of Code Mixed Text}

\author{Ayush Kumar$^{*}$ \qquad    
  Harsh Agarwal$^{*}$ \qquad  
  Keshav Bansal\thanks{\quad Authors equally contributed  to this work.} \qquad
  Ashutosh Modi  \\
{Indian Institute of Technology Kanpur (IITK)} \\
  {\tt \{ayushk,harshaga,keshavb\}@iitk.ac.in}  \\
  {\tt ashutoshm@cse.iitk.ac.in}  \\
}
\date{}
\blfootnote{This work is licensed under a Creative Commons Attribution 4.0 International Licence. Licence details: \url{http://creativecommons.org/licenses/by/4.0/}.}
\begin{document}
\maketitle
\begin{abstract}
Sentiment Analysis of code-mixed text has diversified applications in opinion mining ranging from tagging user reviews to identifying social or political sentiments of a sub-population. In this paper, we present an ensemble architecture of convolutional neural net (CNN)  and self-attention based LSTM for sentiment analysis of code-mixed tweets. While the CNN component helps in the classification of positive and negative tweets, the self-attention based LSTM, helps in the classification of neutral tweets, because of its ability to identify correct sentiment among multiple sentiment bearing units. We achieved F1 scores of 0.707 (ranked $5^{th}$) and 0.725 (ranked $13^{th}$) on Hindi-English (Hinglish) and Spanish-English (Spanglish) datasets, respectively. The submissions for Hinglish and Spanglish tasks were made under the usernames \textit{ayushk} and \textit{harsh$\_$6} respectively.

\end{abstract}



\section{Introduction}
\label{intro}
The research problem of Sentiment Analysis of Code-Mixed Social Media Text appeared as part of the SemEval Shared Challenge 2020 \cite{patwa2020sentimix}. Mixing languages while writing text, also called code-mixing, is a typical pattern observed in almost all forms of communication, including social media text. 
Even though code-mixed languages may contain words from multiple languages, we only focus on bilingual code-mixed languages. 
Two such popular code-mixing styles are Hinglish and Spanglish.
\newline
\indent Sentiment Analysis is a term broadly used to classify states of human affection and emotion. Interpreting code-mixed languages is difficult not only because the sentences may not fit a particular language model, but also because mixed text on social-media usually contains tokens such as hashtags, and usernames.
\newline\indent
In this paper, we present an ensemble of CNN and self-attention based LSTM, utilizing the XLM-R embeddings \cite{xlm_explain}. While CNNs have been used for sentiment analysis before \cite{cnn_cmsa1,cnn_cmsa2}, none of the previous works have used a self-attention based LSTM along with it. We found that while the CNN component worked well for positive and negative tweets, the self-attention component worked better for neutral tweets, necessitating an ensemble of the two.
The implementation of our system is made available via Github\footnote{\url{https://github.com/keshav22bansal/BAKSA_IITK}}.



\section{Related Work}
Performing standard  NLP  tasks on code-mixed data has presented significant challenges. \newcite{vyas-etal-2014-pos} attempted to find methods for POS tagging of code-mixed social media text. 
\newline\indent
Another work by \newcite{joshi-etal-2016-towards} used CNNs to learn subword level embeddings and then utilized these embeddings in a BiLSTM network to learn subword level information from social media text. Subword level representations are particularly important while dealing with noisy texts containing misspellings and punctuations.
However, this work does not capture information related to word-level semantics. This provides further scope to study the impact of word embedding based approaches.
\newline
\indent More recent work by \newcite{lal-etal-2019-de} uses two parallel BiLSTMs, which they call the Collective and Specific Encoder and an additional feature network. This approach combines recurrent neural networks utilizing attention mechanisms, which helps in evaluating the overall sentiment using attention weights when presented with a mixture of local sentiments.

\section{Proposed Approach}
\subsection{Pre Processing}
\begin{figure}[H]
\centering
\begin{minipage}{.48\linewidth}
  \centering
  \includegraphics[scale = 0.15,trim={5.5cm 11cm 5cm 10cm},clip]{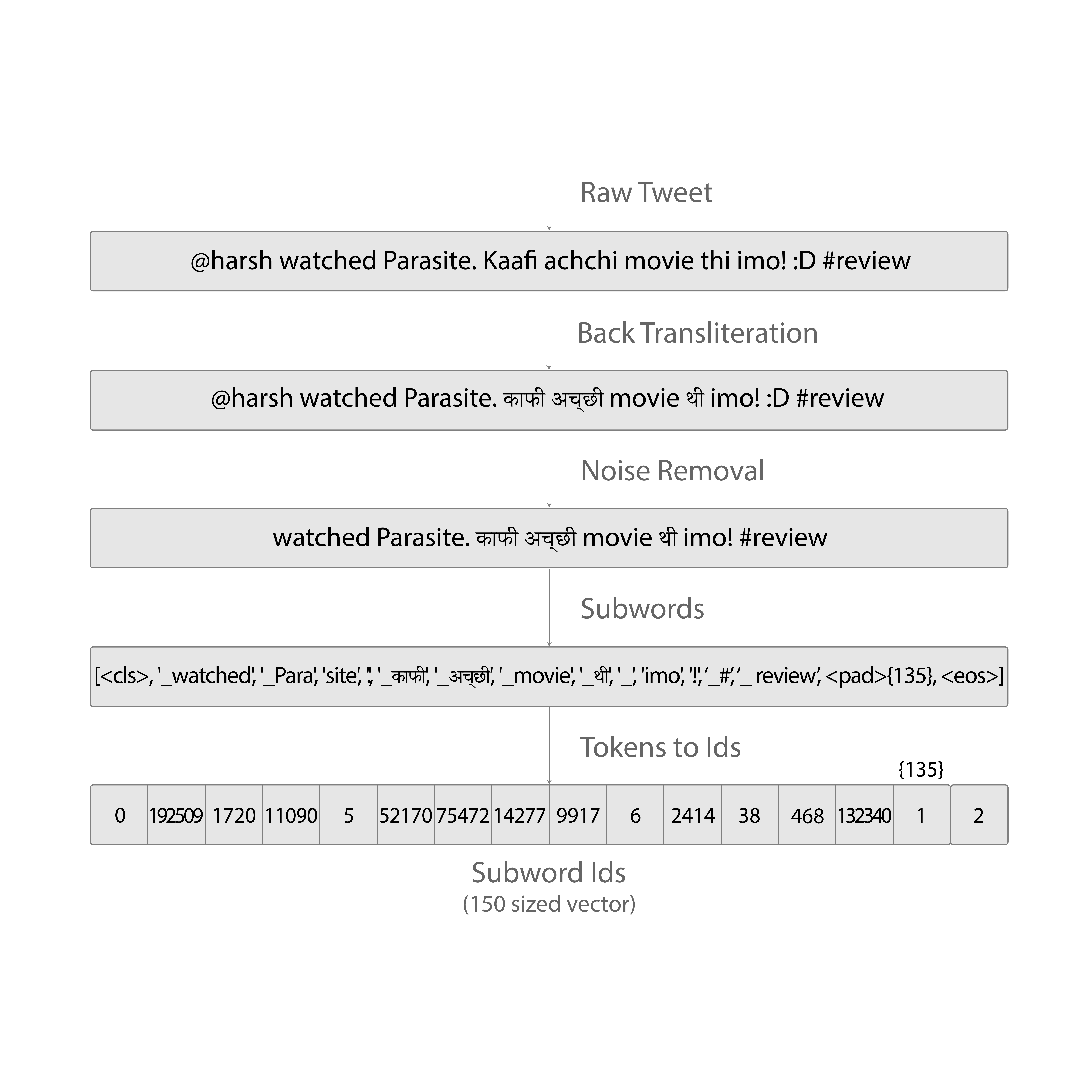}
  \caption*{\hspace{3.5em} Figure 1: Preprocessing Pipeline}
 \label{preprocessing_pipeline}
\end{minipage}
\hfill
\begin{minipage}{.44\linewidth}
  \centering
  \includegraphics[scale=0.15, trim={12cm 10cm 11cm 11cm},clip]{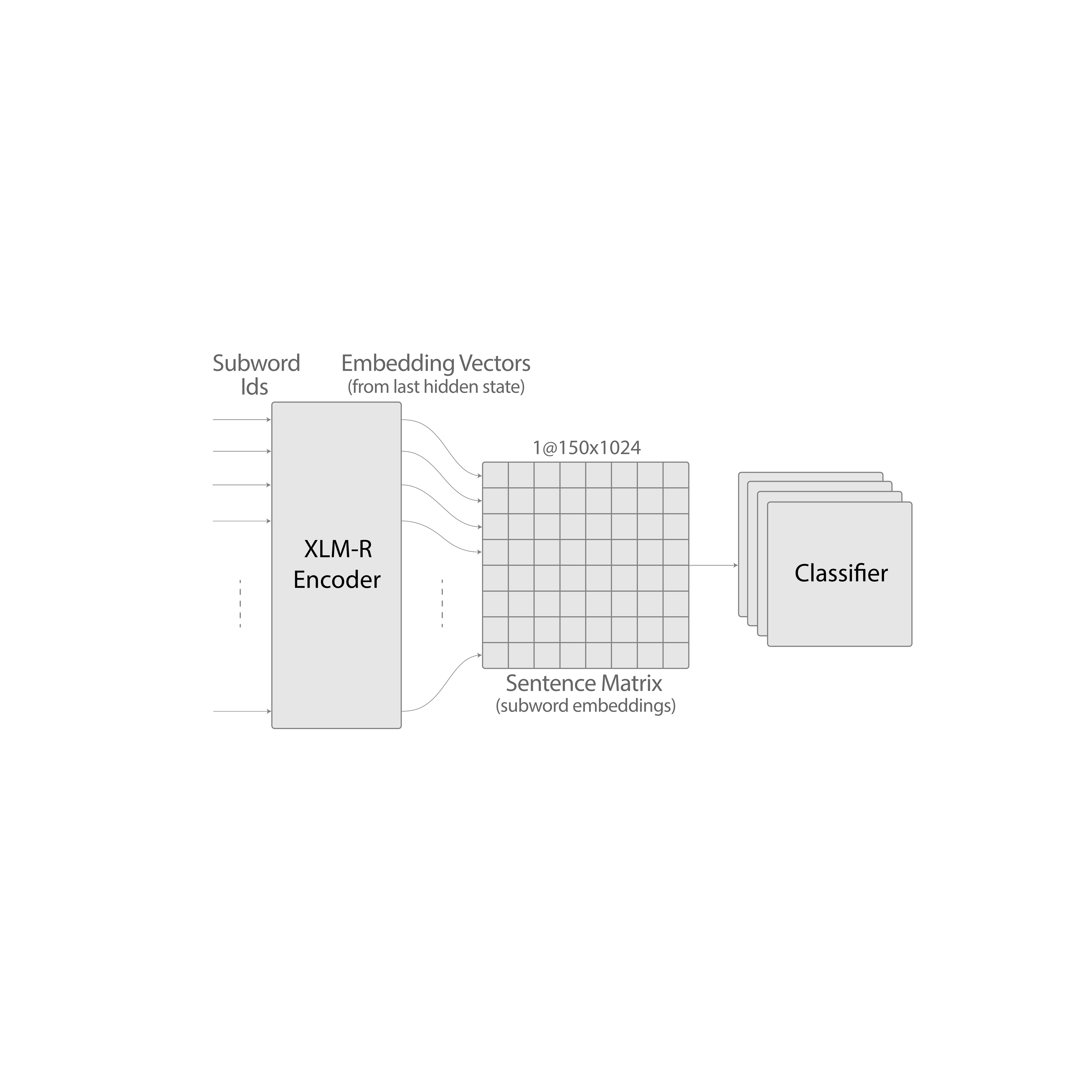}
  \caption*{\hspace{1em} Figure 2: XLM-R Encoder}
  \label{xlmr_encoder}
\end{minipage}
\end{figure}
  
The tweets have been originally provided in the Latin script with their corresponding language tags. Before feeding the tweets to any training stage, they are preprocessed using the following procedure (Figure 1):
\begin{enumerate}
    \item Back-Transliteration: All the words with ``Hindi'' language tags are converted into Devanagari words using phonetic transliteration. Google's Transliteration API\footnote{\url{https://www.google.com/inputtools/services/features/transliteration.html}} was used for this purpose. The words with ``Spanish'' language tags are not transliterated.
    
    \item Noise removal: Usernames (annotated as @username), URLs, and emoticons present in the tweets are removed altogether, while hashtags (annotated as \#hashtag) are left as it is. We also experimented with replacing emoticons by their corresponding textual meaning, but removing them led to better performance.
    
    \item Tokenization: Tweets after noise removal are tokenized into subwords using the XLM-R \cite{xlm_explain} vocabulary and later converted into their corresponding IDs.
\end{enumerate}


\subsection{Embedding layer}
Since our data comprised of code-mixed tweets, it was essential to use a multilingual model. For our proposed architecture, we used the XLM-R embeddings. XLM-R is a transformer-based masked language model trained on one hundred languages, using more than two terabytes of filtered CommonCrawl data \cite{xlm_explain}. 

The subword IDs from the pre-processing stage are fed to the XLM-R encoder. The final hidden state corresponding to each token is used for the classification task as inputs to the proceeding components (See figure 2). The XLM-R encoder is fine-tuned during training to generate better encodings for the code-mixed text.

We also experimented with the Multilingual BERT (henceforth, M-BERT), released by \newcite{DBLP:journals/corr/abs-1810-04805}. We found that XLM-R performed much better than M-BERT for our dataset. 

\subsection{Architecture}
We propose an ensemble model comprising of two main components.
\begin{figure}[H]
\centering
\begin{minipage}{.49\linewidth}
    \centering
  \includegraphics[scale=0.14, trim={6.8cm 16.5cm 5.5cm 16cm},clip]{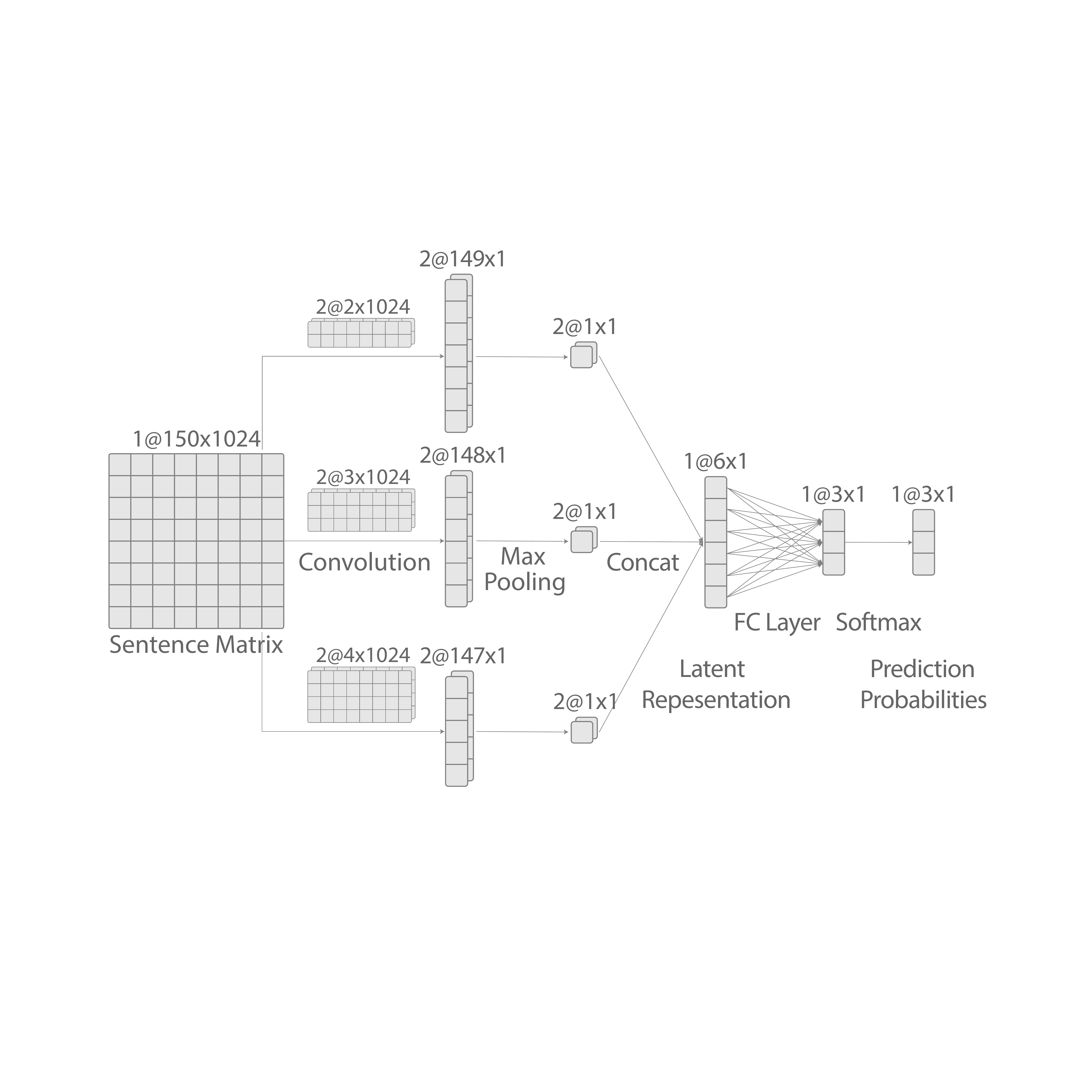}
  \caption*{Figure 3: CNN Classifier}
    \label{img1}
\end{minipage}
\hfill
\begin{minipage}{.50\linewidth}
    \centering
  \includegraphics[scale=0.125, trim={3cm 13cm 2cm 15cm},clip]{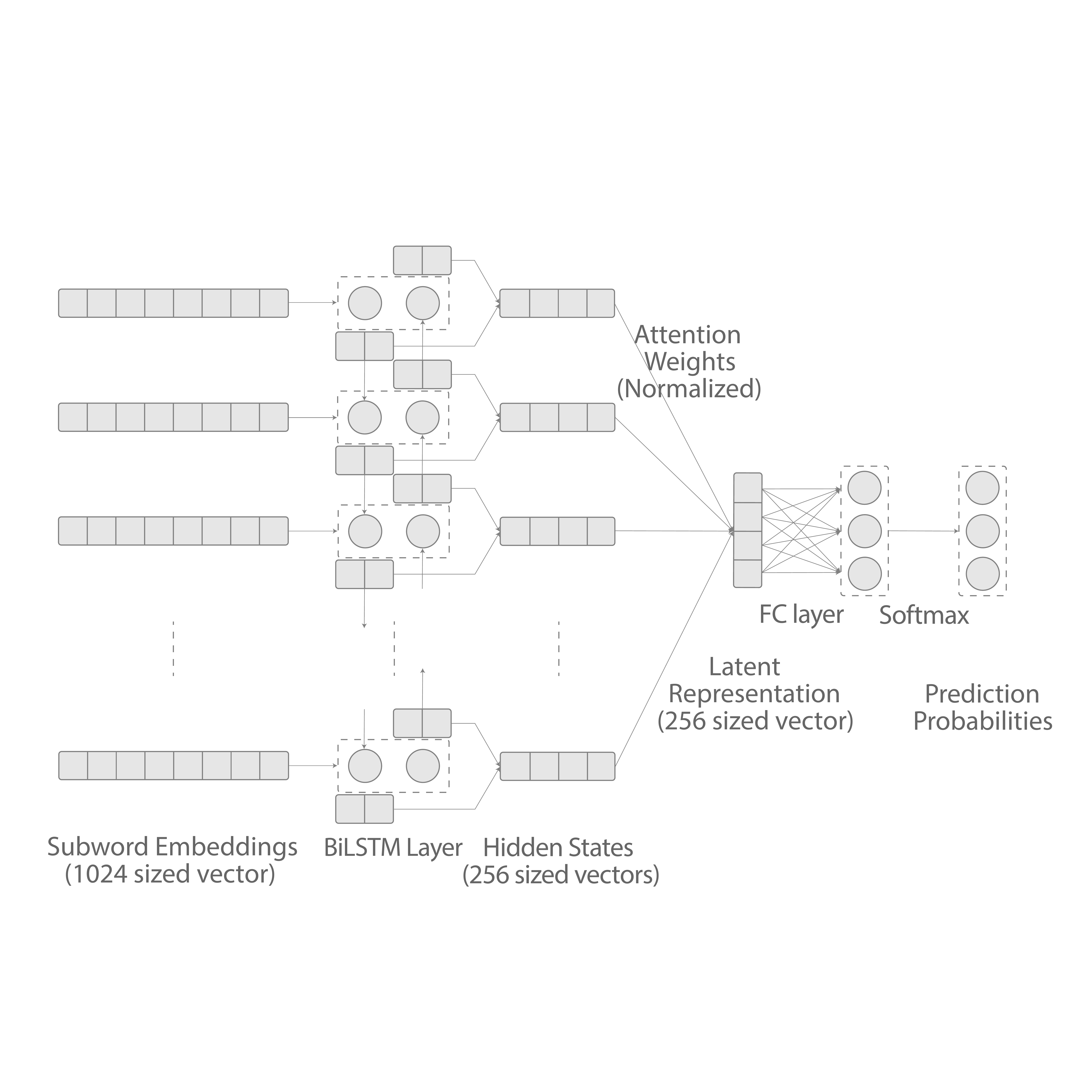}
  \caption*{Figure 4: Self-Attention Classifier}
  \label{self_att}
\end{minipage}
\end{figure}


\subsubsection{CNN Classifier}
The first component is a convolutional neural network \cite{CNN} (henceforth, CNN). CNNs, to some extent, take into account the ordering of the words and the context in which each word appears. 






We generate the required embedding by passing the subword embeddings of a sentence individually into 1-D CNN. We perform a convolution with 3 different filter sizes ($2$, $3$ and $4$), before adding a bias and applying a non-linear RELU activation.

The idea behind using several filter sizes was to capture contexts of varying lengths. The convolution layer is used to extract local features around each word window, while the max-pooling layer is used to extract the essential features in the feature map. XLM-R embeddings are passed through this component and, ultimately, through a softmax function to obtain the predictions of the first component. We call these predictions $p_{CNN}$.

\subsubsection{Self-Attention Classifier}
The second component is a self-attention based classifier (See figure 4).  It helps in choosing the overall sentiment when presented with a mixture of sentiments. We use soft-attention \cite{DBLP:journals/corr/XuBKCCSZB15}, a deterministic, differentiable attention mechanism, where a softmax gives the weights for each subword, and the output of the attention module is a weighted sum of hidden representations at each location.


 
The self-attention component comprises a BiLSTM \cite{lstm} layer, which takes as input the output of the XLM-R encoder. The hidden state obtained from the BiLSTM layer for each subword is used to calculate the attention scores.


Suppose a sequence is given by the subwords $(w_{1},w_{2},...,w_{n})$.  
Let the $i^{th}$ forward hidden state in the BiLSTM be represented by $\overrightarrow{h_i}$ and $i^{th}$ backward hidden state by $\overleftarrow{h_i}$. The combined annotation $k_i$ is obtained by concatenating $\overrightarrow{h_i}$ and $\overleftarrow{h_i}$. We first concatenate the forward and backward hidden states to obtain a combined annotation $(k_{1},k_{2},...,k_{n})$.
\begin{equation}
k_{i} = [\overrightarrow{h_i};\overleftarrow{h_i}]
\end{equation}

The attention mechanism gives a score $e_{i}$ to each subword $i$ in the sentence $S$, as given by (2).

\begin{equation}
e_{i} = {k_i}^{T}{k_{n}}
\end{equation}

Then the attention weight $a_i$ of each $k_i$ is computed by normalizing the attention score $e_i$

\begin{equation}
a_{i} = \frac{{exp(e_i)}}{\sum_{j=1}^{n}{{exp(e_{j})}}}
\end{equation}

We then calculate the sentence latent representation vector $h$ using equation (4)
\begin{equation}
h = \sum_{i=1}^{n}{a_i}\times{k_i}
\end{equation}

The representation is thus a weighted combination of all the hidden states.
The representation vector $h$ is then passed through a fully connected layer followed by a softmax to obtain predictions $p_{att}$.

The predictions from the first and second components are aggregated (See figure 5) using element wise product (denoted by $\circ$) to obtain the final predictions ($p_{final} = p_{CNN} \circ  p_{att}$). We experimented with other aggregating techniques like linearly weighted average, but element-wise product worked out better.
    

\begin{figure}[H]
\begin{minipage}{.49\linewidth}
    \centering
  \includegraphics[scale=0.13, trim={9.5cm 25cm 9cm 25cm},clip]{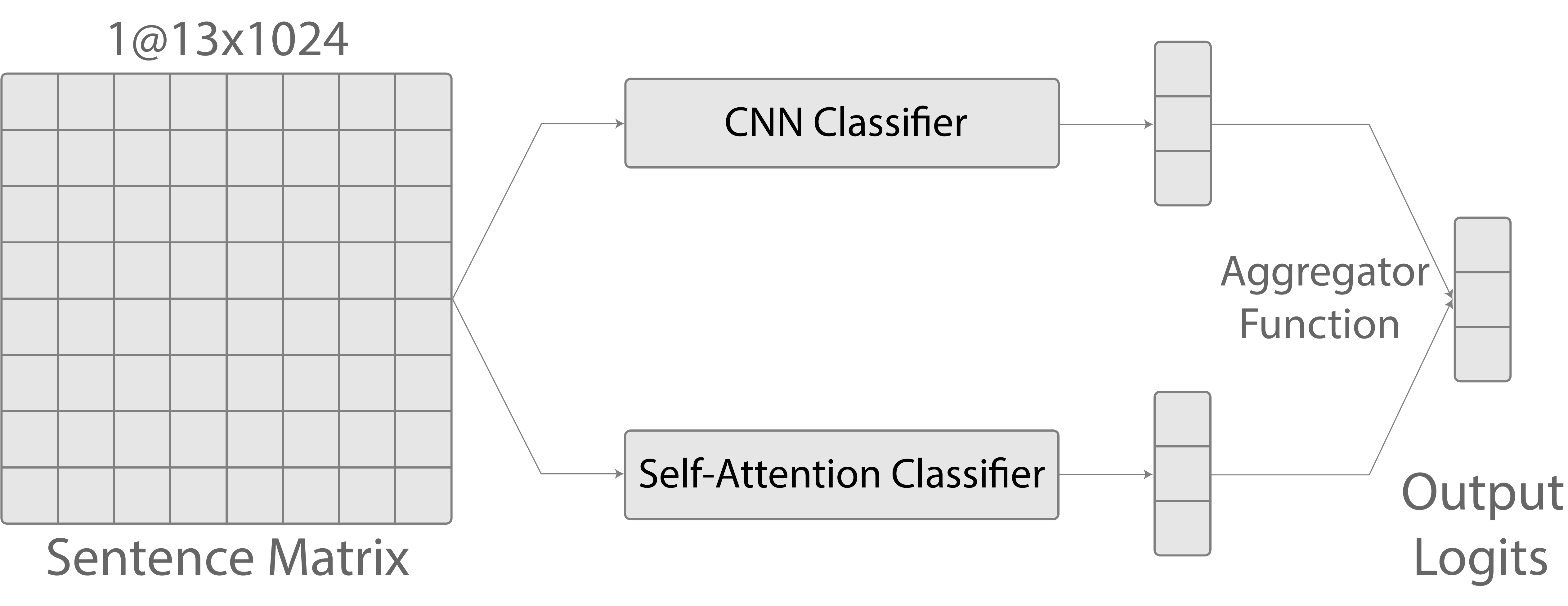}
  \caption*{Figure 5: Ensemble Classifier}
  \label{fig:ensemble}
\end{minipage}
\hfill
\begin{minipage}{.51\linewidth}
\centering
\begin{adjustbox}{width=\columnwidth}{
\begin{tabular}{lllll}
\hline
                                & \textbf{Dataset} & \textbf{Positive} & \textbf{Neutral} & \textbf{Negative} \\ \hline
{\textbf{Train}} & Hinglish         & 5264              & 4634             & 4102              \\ \cline{2-5} 
                                & Spanglish        & 6005              & 3974             & 2023              \\ \hline
{\textbf{Validation}}   & Hinglish         & 982               & 1128             & 890               \\ \cline{2-5} 
                                & Spanglish        & 1498              & 994              & 506               \\ \hline
{\textbf{Test}}   & Hinglish         & 1000               & 1100             & 900               \\  \hline
\end{tabular}
}
\end{adjustbox}
\captionof{table}{Statistics of training and development data}
\label{tabel:stats}

\end{minipage}

\end{figure}
\section{Data Description}
We used the dataset provided by the organizers of Task-9 of SemEval 2020 \cite{patwa2020sentimix} for training both Hinglish and Spanglish models. The data has been annotated semi-automatically.
The statistics of the dataset are shown in Table \ref{tabel:stats}. The dataset for Hinglish is balanced while that of Spanglish is highly unbalanced. For hyperparameter tuning, we used the validation set provided by the organizers.
\section{Experiments and Results}
We first trained a vanilla CNN model on the provided dataset using the XLM-R embeddings. The CNN model seemed to be confused on neutral data points but worked well on positive and negative tweets.
\begin{figure}[H]
\captionsetup{justification=centering}
\centering
\begin{minipage}{.48\linewidth}
    \centering
  \includegraphics[scale=0.09]{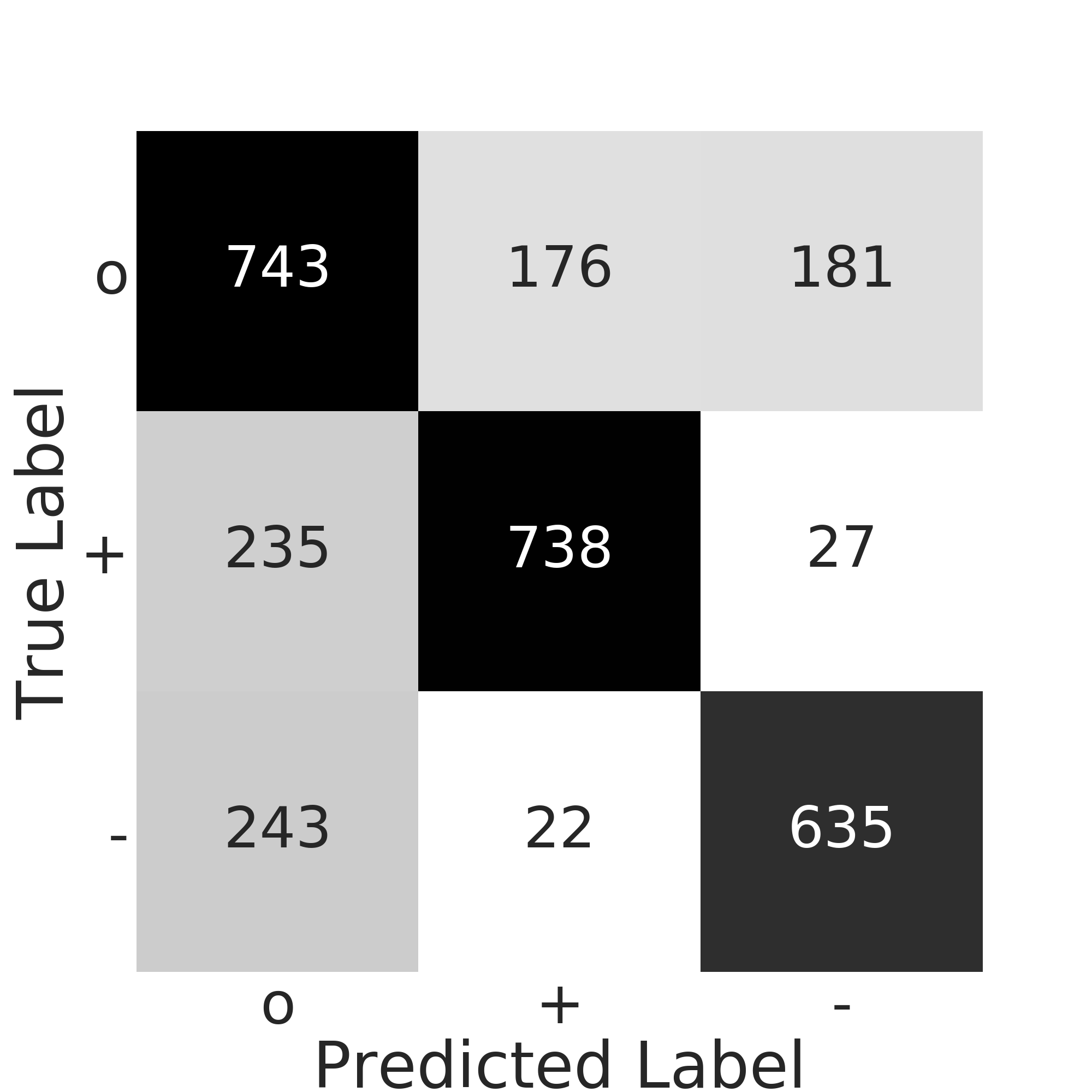}
  \caption*{Figure 6: Confusion matrix for Ensemble on Hinglish test data}
    \label{img:conf_hinglish}
\end{minipage}
\hfill
\begin{minipage}{.51\linewidth}
    \centering
  \includegraphics[scale=0.09, trim={0cm 0cm 0cm 0cm},clip]{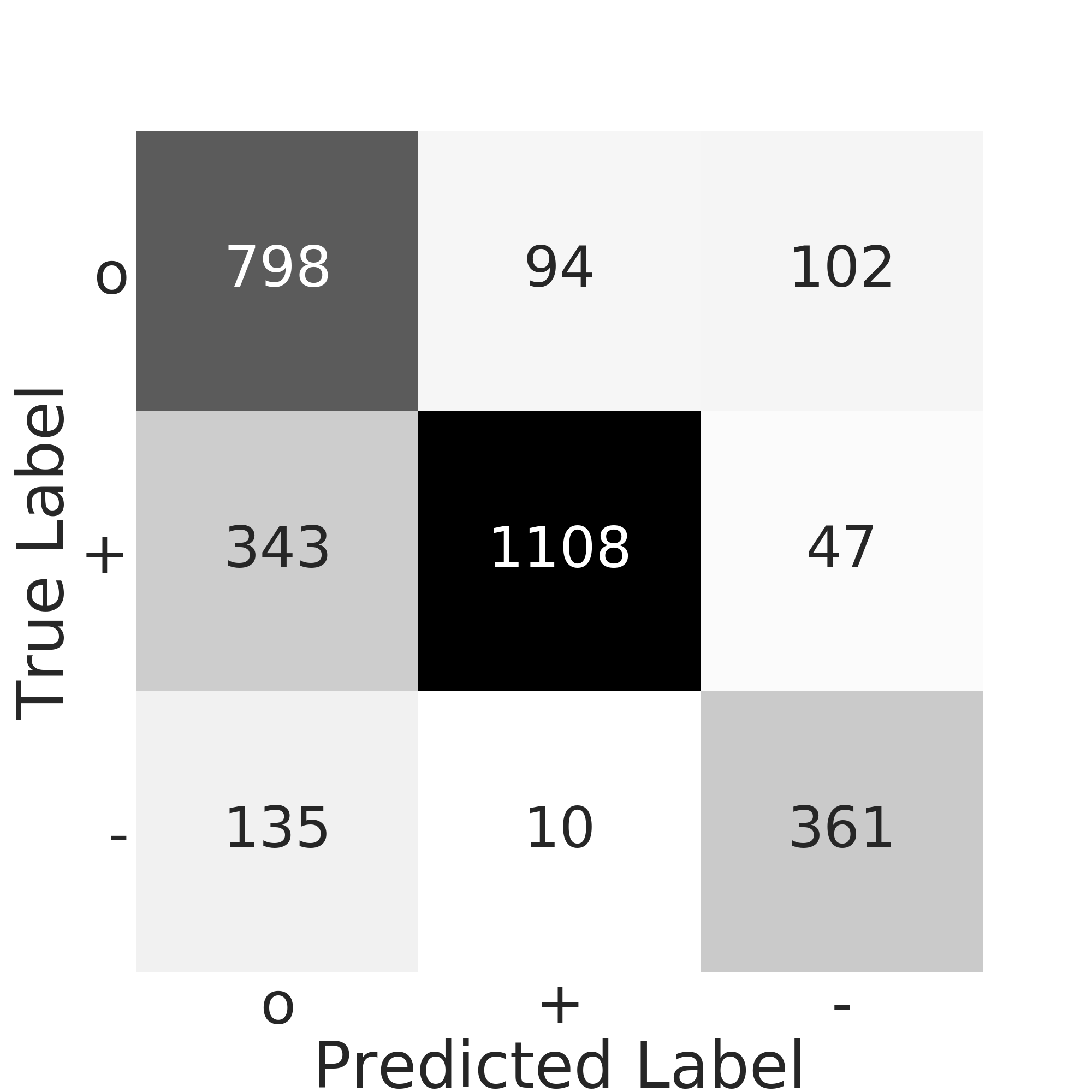}
  \caption*{Figure 7: Confusion matrix for Ensemble on Spanglish validation data
  \footnotemark
}
  \label{img:cconf_spanglish}
\end{minipage}

\end{figure}
 \footnotetext{{Validation data was used for constructing the confusion matrix for spanglish as true labels for test data were not available}}




\indent The self-attention model outperforms the previous model on neutral data points though it performs worse on the positive and negative samples. The good performance on neutrals can be attributed to the fact that neutral tweets may contain multiple sentiment bearing units which the model is capable of handling.

\indent Combining the results of CNN with those of the Self-Attention model was the primary motivation for using an ensemble of the two. The ensemble outperforms all our previous models, achieving a recall of 0.705 with an F1-score of 0.707 on the Hinglish test dataset and a recall of 0.696 with an F1-score of 0.725 on the Spanglish test dataset (See table \ref{tab:system_scores}). The confusion matrices for the ensemble on both datasets are shown in figure 6 and 7 (o : neutral, + : positive, - : negative). Our team was ranked $5^{th}$ among 62 teams in Hinglish and $13^{th}$ among 29 teams in Spanglish.

\begin{table}[t]
\centering
\begin{adjustbox}{width=0.5\columnwidth}{
\begin{tabular}{@{}ccccccc@{}}
\toprule
 & \multicolumn{4}{c}{\textbf{F1}}                       & \textbf{Macro}     & \textbf{Macro}  \\ \cmidrule(lr){2-5}
                          & \textbf{o} & \textbf{+} & \textbf{-} & \textbf{Macro} & \textbf{Precision} & \textbf{Recall} \\ \hline
\textbf{Hinglish}          & 0.640       & 0.762       & 0.729       & 0.707           & 0.712               & 0.705            \\
\textbf{Spanglish}         & 0.135      & 0.825       & 0.375      & 0.725           & 0.763               & 0.696            \\ \bottomrule
\end{tabular}
}\end{adjustbox}
\caption{Performance of Ensemble system on Hinglish and Spanglish test datasets}
\label{tab:system_scores}
\end{table}

\section{Analysis}
\subsection{Visualization of the individual components}
To visualize the sentence embeddings learned by the model for the Hinglish test dataset, we projected the sentence vectors obtained before the final fully connected layer onto a lower-dimensional subspace using the t-SNE algorithm \cite{vanDerMaaten2008} for the two components (See figure 8).

For CNN, the positive and negative tweets seem to form two distinct clusters, while the neutral tweets are scattered among them. In contrast, for the self-attention component, neutrals seem to form a distinct cluster, while the positive and negative classes are partially dispersed in a wide region. Thus, the two components, in a way, complement each other for better predictions over all the three classes.



\begin{figure}[htbp]
\centering
\includegraphics[scale=0.34, trim={0cm 1.1cm 0cm 1cm},clip]{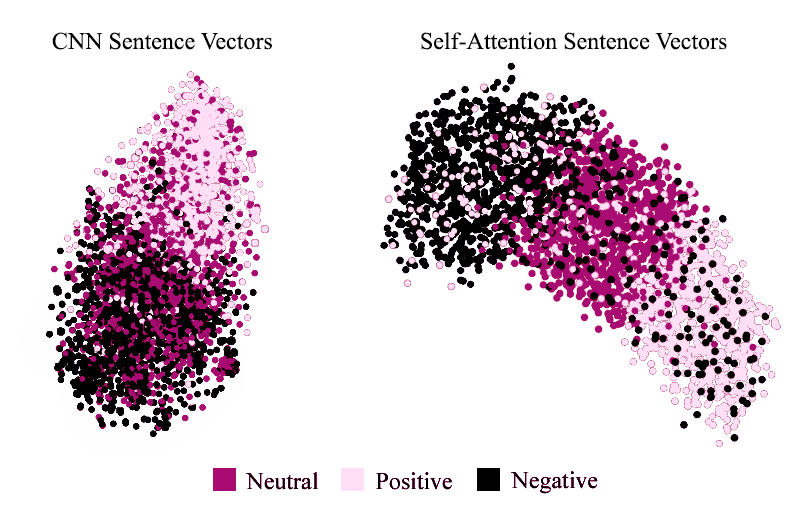}
\caption*{Figure 8: Visualisation of CNN and Self-Attention Sentence Vectors}
\label{fig:sentence_vis}
\end{figure}



\subsection{Error Analysis}
Most of the misclassifications were made by our model on the following three types of tweets - 
\begin{enumerate}
    \item Neutral - Despite the improvement due to the self-attention classifier, the performance on neutral tweets still lags much behind positive and negative tweets.
    \item Sarcastic - Sarcasm is the use of irony to mock or convey contempt. Tweets such as \textbf{\textit{Best wishes to pseudo atheist In new country in advance. Bon voyage}} are challenging to classify due to their hidden context and are falsely predicted as positive by our model.
    \item Mildly negative - Due to exorbitant amount of abusive tweets in the data, some mildly negative ones like \textbf{\textit{South africa team bekar h jab tak ushme ABD villers na ho}} are falsely predicted as neutral. 
\end{enumerate}


\section{Conclusion}
For our system, we use an ensemble of CNN and Self Attention architectures with XLM-R multilingual embeddings.
We analyze which models work better for different classes of tweets. Our self-attention system helps in better classification of neutral tweets, which are difficult to classify due to multiple sentiment bearing units. Creating an ensemble with CNN helps in better classification of all the three classes. We also visualize how our model performs on different classes of tweets using the t-SNE algorithm. Our results show an improvement over some of the previous works in this field.

\bibliographystyle{coling}
\bibliography{semeval2020}

\end{document}